\documentclass[pdflatex,sn-mathphys-num]{sn-jnl}

\usepackage{graphicx}%
\usepackage{multirow}%
\usepackage{amsmath,amssymb,amsfonts}%
\usepackage{amsthm}%
\usepackage{mathrsfs}%
\usepackage[title]{appendix}%
\usepackage{xcolor}%
\usepackage{textcomp}%
\usepackage{manyfoot}%
\usepackage{booktabs}%
\usepackage{algorithm}%
\usepackage{algorithmicx}%
\usepackage{algpseudocode}%
\usepackage{listings}%
\usepackage{tabularx}

\theoremstyle{thmstyleone}%
\theoremstyle{thmstyletwo}%

\theoremstyle{thmstylethree}%

\begin{document}

\title{Dynamic Knowledge Fusion for Multi-Domain Dialogue State Tracking}

\author[1,2]{\fnm{Haoxiang} \sur{Su}} \equalcont{These authors contributed equally to this work.}

\author[2]{\fnm{Ruiyu} \sur{Fang}} \equalcont{These authors contributed equally to this work.}


\author[1]{\fnm{Liting} \sur{Jiang}} 

\author[2]{\fnm{Xiaomeng} \sur{Huang}}





\author*[2]{\fnm{Shuangyong} \sur{Song}}\email{songshy@chinatelecom.cn}



\affil[1]{School of Computer Science and Technology, Xinjiang University}

\affil[2]{Institute of Artificial Intelligence (TeleAI), China Telecom Corp Ltd}



\abstract{The performance of task-oriented dialogue models is strongly tied to how well they track dialogue states, which records and updates user information across multi-turn interactions. However, current multi-domain DST encounters two key challenges: the difficulty of effectively modeling dialogue history and the limited availability of annotated data, both of which hinder model performance. To tackle the aforementioned problems, we develop a dynamic knowledge fusion framework applicable to multi-domain DST. The model operates in two stages: first, an encoder-only network trained with contrastive learning encodes dialogue history and candidate slots, selecting relevant slots based on correlation scores; second, dynamic knowledge fusion leverages the structured information of selected slots as contextual prompts to enhance the accuracy and consistency of dialogue state tracking. This design enables more accurate integration of dialogue context and domain knowledge. Results obtained from multi-domain dialogue benchmarks indicate that our method notably improves both tracking accuracy and generalization, validating its capability in handling complex dialogue scenarios.}




\maketitle

\section{Introduction}\label{sec1}
As artificial intelligence technologies continue to advance, The adoption of dialogue systems has expanded rapidly, supporting a variety of real-world applications that span clinical consultation platforms, and digitally empowered governmental services\cite{wu2025multi},\cite{song2017intension}. Unlike traditional single-task dialogue systems, real-world interactions often involve the integration of information from multiple domains. For instance, in a single conversation, a user may inquire about hotel reservations, flight arrangements, and restaurant recommendations simultaneously. This requires the system to flexibly switch across domains and accurately track user intentions. Against this backdrop, Dialogue State Tracking (DST) has emerged as an essential module within task-oriented dialogue systems\cite{su2023scalable},\cite{xie2022correctable},\cite{song2024improving},\cite{song2024graph},\cite{wu2025int},\cite{su2025raicl},\cite{xie2025mitigating},\cite{pang2022mfdg},\cite{xu2023improving},\cite{xiong2025tablezoomer}. DST mainly aims to capture the semantics of user inputs and track information over multiple dialogue turns, thereby constructing an accurate dialogue state that facilitates downstream decision-making and response generation. However, in multi-domain settings, DST not only needs to model complex dialogue history and contextual information but also faces challenges caused by data scarcity and insufficient domain knowledge, placing higher demands on the model’s generalization capability.

Existing research has shown that both structured and unstructured knowledge play essential roles in DST. Schemas and ontologies are typical forms of structured knowledge. Schema knowledge can be viewed as system metadata, defining the supported domains, slots, and their semantic ranges, which helps the model better understand user intentions. Ontology knowledge, on the other hand, specifies slot types and their possible values, constraining and guiding the interpretation of complex semantic logic. At the same time, unstructured knowledge such as large-scale text corpora provides rich background information that enhances the robustness of dialogue understanding. Nevertheless, pretrained language models typically lack domain-specific knowledge, which reduces their capability when handling multi-domain scenarios. In response to this challenge, numerous studies have explored the incorporation of external knowledge sources to improve the effectiveness of DST models. One line of work encodes schema knowledge directly into the model to improve context modeling\cite{hosseini2020simple},\cite{madotto2020learning}; another reformulates DST as a question answering (QA) task\cite{zhou2019multi},\cite{wu2020tod},\cite{lee2021dialogue},\cite{ren2018towards},\cite{rastogi2020towards},\cite{du2021qa}, thereby introducing ontology knowledge more naturally; yet another concatenates all slot and slot-value information with dialogue context to provide comprehensive knowledge support. While these methods achieve performance gains to some extent, they also exhibit limitations\cite{zhao2022description}. First, directly encoding schema or ontology information is often inefficient in multi-domain settings and difficult to scale. Second, reformulating DST as QA requires querying slot values one by one, which increses computational cost and limits scalability. Third, simply concatenating all slots and values may lead to “attention dilution”\cite{fan2021augmenting}, impairing the model’s capacity to identify and attend to the most essential signals, ultimately reducing performance. Thus, an open challenge in multi-domain DST is how to efficiently utilize structured knowledge while maintaining scalability\cite{hong2023knowledge}.

To tackle these issues, we proposes a dynamic knowledge fusion model for multi-domain DST (DKF-DST). The central concept behind the model is to incorporate structured knowledge dynamically through a relevant slot selection mechanism, thereby avoiding the introduction of invalid or redundant information. Specifically, the model first identifies slots most relevant to the current dialogue context via relevance computation. It then dynamically incorporates the corresponding schema and ontology knowledge as prompts to guide the encoding process, enabling more precise modeling of user intentions and dialogue states. Compared to traditional approaches, this method improves the alignment between dialogue context and outside knowledge sources, mitigates the issue of attention dispersion, and ensures scalability in multi-domain environments.

The main contributions of the model are summarized below:
\begin{itemize}
   \item We introduce dynamic knowledge fusion mechanism that significantly enhances the precision and generalization capability in multi-domain dialogue state tracking (DST), providing robust technical support for real-world dialogue system deployment.

    \item We introduce a novel perspective for combining structured knowledge with pretrained language models, broadening research directions in knowledge-augmented dialogue modeling.

    \item Experiments on multi-domain dialogue datasets show that our model surpasses existing baselines, confirming the effectiveness and feasibility of dynamic knowledge fusion for enhancing multi-domain DST performance.

\end{itemize}

\section{Related work}
The dialog state tracking models primarily relied on rules and heuristics, using manually designed rules to manage dialogue states. However, these methods performed poorly in complex dialogue scenarios. Rule-based DST approaches identify dialogue states through predefined rules and templates \cite{goddeau1996form}. While these methods offer strong interpretability, they require extensive manual effort and domain expertise, making them difficult to scale to complex dialogues. Consequently, they suffer from limited generalization, high error rates, and poor adaptability across domains \cite{williams2014web}. In recent years, neural network–driven dialogue state tracking (DST) methods have utilized deep learning models to model and predict dialogue states. According to how slot values are predicted, these methods can be categorized into two types \cite{zhang2019find}: one is the classification-based approach, which selects slot values from a predefined set \cite{mrkvsic2016neural},\cite{su2024domain}; the other is the generation or extraction approach, which either generates slot values directly using the model \cite{wu2019transferable},\cite{ren2019scalable} or extracts segments from the dialogue context to represent slot information \cite{gao2019dialog},\cite{zhou2019multi}.

With the rise and evolution of pre-trained language models in recent years\cite{yang2025code}, it has become increasingly clear that slots in dialogue state tracking (DST) models are not independent but interrelated. Recent advancements in large language models (LLMs), such as the TeleChat series \cite{he2024telechat},\cite{wang2024telechat},\cite{wang2025technical},\cite{liu2025training}, have demonstrated strong capabilities in understanding complex dialogue contexts and relationships between semantic slots. For example, the slot values for "hotel-star rating" and "hotel-price range" are correlated, making slot relationship modeling crucial for accurate DST. Ye et al. \cite{ye2021slot} introduced an approach to explicitly model these relationships using self-attention networks. Following a similar idea, Lin et al. \cite{lin2021knowledge} adopted a combined architecture that merges the GPT-2 model for sequential value prediction using a graph attention network (GAT) to capture and represent relationships among slots and their values. Another line of research leverages the hierarchical structure of domain ontologies and GATs to fuse data from history of dialogue and slot description graphs \cite{li2021generation}. Feng et al. \cite{feng2022dynamic} improved the framework by incorporating a mechanism that continuously updates inter-slot dependencies within the schema graph as the dialogue progresses, resulting in more precise modeling of slot semantics in DST. Su et al. \cite{su2023schema} introduce a graph-based framework for slot relationship modeling, combining dialogue schemas with domain-specific prompts, achieving superior performance over alternative multi-domain DST methods and requiring a comparable or even smaller number of trainable parameters. The continuous evolution of LLM architectures, emerging paradigms such as AI Flow \cite{an2025ai},\cite{shao2025ai}, offer promising directions for enhancing DST through improved contextual understanding and efficient deployment. 

to address the challenges of difficult historical modeling and scarce annotated data in multi-domain dialogue state tracking (DST). Compared to previous methods, the innovations of this work are mainly reflected in: (1) Proposing a two-stage architecture, where a contrastive learning encoder explicitly selects relevant slots, and then their structural information is fused as dynamic knowledge prompts, achieving precise integration of dialogue context and domain knowledge;
(2) Introducing a dynamic knowledge injection mechanism that can adaptively update prompts with the dialogue progress, outperforming static ontology or fixed prompt methods; (3) Enhancing the model's generalization capability with limited annotated data through contrastive learning. Experiments show that this method significantly improves the tracking accuracy and robustness of multi-domain dialogue states.

\section{Methodology}

\begin{figure}[h]
\centering
\includegraphics[width=1.0\textwidth]{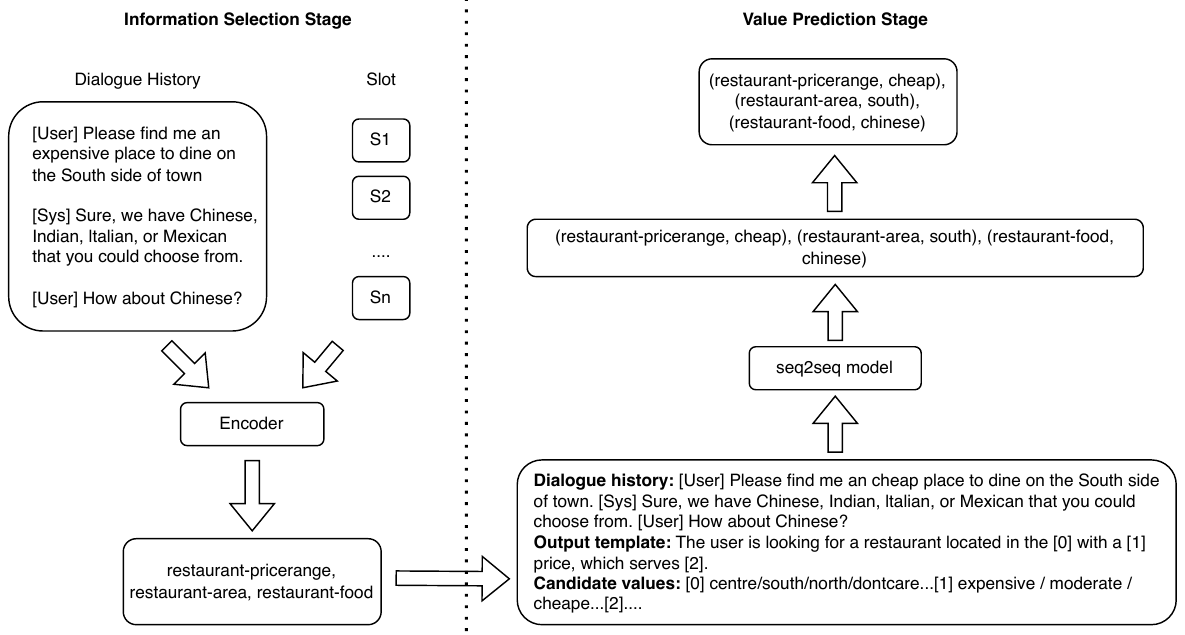}
\caption{The Model Architecture of DKF-DST}\label{model}
\end{figure}

A comprehensive outline of the model’s structural design is presented in this section. We put forward DKF-DST, a novel dialogue state tracking model that leverages dynamic knowledge fusion to operate robustly across various domains. The proposed model adopts a two-stage architecture, where each stage is designed to perform distinct tasks and functions. As shown in Figure \ref{model}, the model comprises two key stages: The first stage focuses on selecting the slot information that requires attention in the dialogue, specifically those slots with non-empty values in the labels. The second stage integrates relevant slot ontology knowledge, receives the dialogue history and the output template to be filled, and allows the model to produce and output the dialogue state in natural language form.

\subsection{Information Selection Based on Contrastive Learning}
First, in the first stage of the model, as shown in the left part of Figure \ref{model}, it is necessary to determine the information that needs to be focused on and processed in the current dialogue — that is, to select the slots with non-empty values and identify the most critical information in the conversation, thereby providing guidance for the subsequent dynamic knowledge fusion.

Regarding the information extraction approach\cite{li2025mr}, in studies that reformulate the dialogue state tracking (DST) task as a question-answering problem, each slot is queried individually, meaning that only one slot’s information is involved per query. However, for DKF-DST, it is necessary to handle multi-domain dialogues that involve multiple slots simultaneously, which means dealing with multiple slot-value pairs and their corresponding dialogue contexts. Therefore, when processing such complex dialogues, it is crucial to accurately determine which slots need attention. Traditional knowledge retrieval methods are also unsuitable in this context, as they rely on retrieving content with representations similar to the target itself. In contrast, for the slot determination task, the dialogue context and target slot do not necessarily share similar representations. Likewise, traditional non-parametric retrieval methods such as TF-IDF and BM25 \cite{robertson2009probabilistic}, which depend on lexical overlap, are also inappropriate for this stage, since the word overlap between dialogue utterances and slot names is often unreliable. For instance, in DST datasets, it is common that slots from different domains share the same value (e.g., the term “cheap” could correspond to the slot “hotel-pricerange” or “restaurant-pricerange”). This overlap further increases the difficulty of using word-level matching.

Considering these limitations, to narrow the representation gap between relevant slots and dialogue history and to improve the selection of relevant information, this study adopts a contrastive learning-based method\cite{xiong2024dual}. Specifically, a single encoder-only model serves to encode both slot as well as dialogue history representations, and the encoder is optimized by minimizing a binary cross-entropy–based contrastive loss. The RoBERTa \cite{liu2019roberta} is selected as the encoder backbone to encode dialogue history and slot representations.

The model is based on RoBERTa, whose structure is largely consistent with BERT and consists of multiple Transformer blocks. By exploiting the bidirectional modeling capability of the Transformer encoder, it transforms textual input into high-dimensional vector representations, in doing so, the model gains the ability to operate across various linguistic tasks, from labeling text and identifying key terms to addressing user inquiries. The core idea is to pretrain the Transformer encoder using large-scale text data and then fine-tune it.
Within each layer block, the multi-head self attention operation and the position-wise feed forward transformation are jointly employed, and both are stabilized through residual connection schemes: the attention component captures contextual information, and the feed-forward layer re-embeds the resulting vectors into a new representation space. More specifically, the model’s structure can be represented as follows:
\begin{equation}
h_{l} = TransformerBlock (h_{l-1}),
\end{equation}
where, $h_l$ denotes the output vector of the $l$-th layer, $h_{l-1}$ denotes the output vector of the ($l-1$)-th layer.

The model’s input is constructed from token, positional, and segment embeddings. While token embeddings encode the text itself, positional embeddings record the location of each token, and segment embeddings represent the boundaries between different sentences in the input.

Central to the model is the Transformer’s self attention mechanism, complemented by a feed forward neural network. Self-attention assesses pairwise relationships among the Q, K, and V vectors, and combines them through a weighted summation to produce the output representation. The computation is expressed as follows:
\begin{equation}
Attention(Q,K,V)=Softmax(\frac{QK^{t}}{\sqrt{d_{k}} } )V,
\end{equation}
where $d_{k}$ indicates the size of the attention head. In this process, the model measures the similarity between the query and key vectors, applies a softmax operation to obtain normalized weights, and then uses these weights to combine the value vectors into the final output.

Secondly, the model utilizes a feed-forward neural network, which applies a nonlinear transformation to the output generated by the self-attention layer. This component is made up of two linear transformations combined with an activation function, and its computation is formulated as follows:
\begin{equation}
FFN(x)=ReLU(xW_{1}+b_{1})W_{2}+b_{2},
\end{equation}
the multilayer perceptron structure performs feature mapping and transformation, which helps enhance the model’s representational capability.

Specifically, the RoBERTa model introduces several key improvements over BERT:
(1) Training data enhancement: RoBERTa is pretrained on a larger and more diverse dataset, including webpages, forums, books, and news articles from the Internet. In addition, it uses longer text sequences during training, allowing it to better capture contextual dependencies.
(2) Training strategy optimization: RoBERTa adopts longer training durations, smaller batch sizes, and higher learning rates, which enhance the model’s robustness and overall performance.
(3) Improved masked language model (MLM) objective: Unlike original BERT, RoBERTa employs a more rigorous masking strategy. Specifically, it replaces all tokens in the input with “[MASK]” and subsequently expects the model to accurately generate the corresponding tokens, thereby making fuller use of the training data.
(4) Larger parameter scale and deeper network structure: RoBERTa increases both the parameter count and the network depth during pre-training, further improving model performance.

The choice of RoBERTa as the encoder foundation is based on its outstanding performance, strong generalization ability, superior handling of long texts, and high scalability. RoBERTa provides powerful support for slot ranking and attention recognition tasks. By employing optimization techniques such as larger batch sizes, longer training durations, and dynamic masking strategies, it enhances both model performance and effectiveness. These improvements enable RoBERTa to better capture the semantic and syntactic structures within text, facilitating accurate understanding of dialogue history and relationships among slots, thereby enabling effective slot ranking and selection.

After loading the RoBERTa model, the training process involves encoding multiple slots and dialogue histories. The model is trained based on the concept of contrastive learning, where the encoder is optimized by minimizing the binary cross-entropy loss through a contrastive objective:
\begin{equation}
L_{con} = -\sum_{n}^{i=1} \alpha_{i}\cdot log(sim(Enc(D),Enc(s_{i})))+(1-\alpha_{i})\cdot log(1-sim(Enc(D),Enc(s_{i}))),
\end{equation}
where, $D$ denotes the dialogue history, and $s_i$ represents a slot. When $s_i$ is a slot that needs to be attended to (i.e., the value of the slot in the reference dialogue state is non-empty), $\alpha_i$ is set to 1; when $s_i$ is an irrelevant slot, $\alpha_i$ is set to 0. In this way, the model can reduce the representation distance between the dialogue history and its corresponding relevant slots. The function \text{sim}() is computed as the dot product of the first token representations of two texts, which serves as the relevance score. This training enables the encoder to effectively link slots with the surrounding dialogue, capturing their semantic interactions more precisely. Based on the relevance score between the dialogue context and each slot, a threshold $\delta$ is set as a hyperparameter. Slots with relevance scores higher than the threshold $\delta$ are ultimately identified as the slots that need to be attended to.

\subsection{Dynamic Knowledge Fusion for State Prediction}

The second stage involves integrating relevant slot ontology knowledge, dialogue history, and a fill-in template so that the model can generate dialogue states in natural language form. Through this process, the “dynamic fusion” of the model is primarily reflected in its ability to incorporate only the relevant information and knowledge—based on the slots identified in the first stage—into the input dynamically, thereby achieving more accurate dialogue state predictions. To accomplish this, multiple slots are transformed into fillable natural language summary templates, and the ontology knowledge corresponding to the relevant slots is concatenated into the input to enable dynamic knowledge fusion. Notably, this stage employs the T5 model, a large-scale pretrained sequence-to-sequence language model.

In the second stage, a Seq2Seq model is employed for modeling. First, the Seq2Seq model is a flexible and general-purpose architecture capable of handling language instructions in various formats. Second, it has been proven effective for applications in dialogue state tracking. Third, as a universal model architecture, the Seq2Seq model can be conveniently initialized from publicly available pretrained checkpoints. Specifically, the DKF-DST model adopts the T5 model.

The proposed model leverages T5, a Transformer-based Seq2Seq architecture that integrates diverse NLP tasks into one unified training framework. It maintains the standard Transformer encoder–decoder design, where the encoder produces contextualized representations of the input text and the decoder generates the corresponding target sequence.

During training, the model follows a text-to-text paradigm, transforming various NLP tasks into a format where input text is converted into corresponding output text for training. When fine-tuning for downstream tasks, a prefix similar to an explicit prompt is prepended to the input sequence to indicate the type of task that T5 needs to solve. This prefix is not merely a label or a classification indicator like [CLS]; instead, it captures the essence of the task that the Transformer needs to perform. The prefixes include:
(1) translate English to German: + sequence – Translation task
(2) cola sentence: + sequence – CoLA corpus, fine-tuning the BERT model
(3) stsb sentence 1: + sequence – Semantic Textual Similarity benchmark. Tasks such as natural language inference and entailment are similar
(4) summarize + sequence – Text summarization task

In this way, a unified format covering a wide spectrum of NLP tasks is obtained, capable of handling various text-to-text tasks. The T5 model is used because of its powerful modeling capabilities, strong performance across diverse tasks, rich language representation ability, flexibility, and scalability, as well as its broad recognition and application within the natural language processing community.

Moreover, T5 has been proven in cutting-edge research and practice to achieve excellent performance on numerous tasks. Thanks to its text-to-text training framework, T5 can perform end-to-end sequence-to-sequence transformations without requiring additional processing or decoding steps, which helps simplify the model architecture and improve efficiency.

The model input integrates the idea of prompt learning, inspired by previous work \cite{yu2022knowledge}, \cite{zhao2022description}, and further extends these approaches. Figure \ref{model} (right) shows that the model takes the following components as input:

Dialogue history: To encode the dialog context thoroughly and capture complete contextual information, the full dialogue history is incorporated, including the content of preceding dialogue turns. User utterances and system responses are explicitly differentiated using the tags [User] and [Sys].

Output template prompt: Based on the relevant slots selected in the previous stage, a dynamic output template is constructed with reference to the dialogue state transformation templates provided in prior studies. The output template uses masks (e.g., [0], [1], etc.) to mark the positions where slot values need to be filled in.

Candidate values: Dynamic slot candidate values are incorporated into the input prompt message, corresponding to each masked slot position. These candidate values are derived from the ontology knowledge contained in the dataset.

The output template is constructed using the summarization template method proposed in the DS2 model \cite{shin2022dialogue}, where templates are generated based on the slots predicted in the previous stage. Given a dialogue state, the corresponding summary is generated in a hierarchical manner following the predefined template.
Assume the current domain contains $m$ slots, denoted as $k_1$, …, $k_m$. For each slot $k_i$, a phrase template $p_i$ is defined, functioning as a mapping that receives a value string and outputs a corresponding phrase. For example, the slot “attraction-area” corresponds to the template “located in …”. The slots predicted in the previous stage are matched with their respective phrase templates to obtain a collection of phrases, which are then combined and appended to the domain-related prefix sentence, with masks added to indicate missing values — for example, “The user is looking for [0]”. For instance, if the slots predicted in the previous stage are “taxi-departure” and “taxi-destination,” then in the second stage, the embedded output template prompt in the model input would be:
“The user is looking for a taxi from [0] to [1].”
In the candidate values section of the input prompt, ontology knowledge is incorporated by appending the candidate values corresponding to the masked slots “taxi-departure” and “taxi-destination” after their respective mask positions.

The model output is generated by training T5 to fill masked slots in a predefined template, thereby producing a coherent natural-language summary of the dialogue state. The final dialogue state is obtained by reversing the template to retrieve the relevant slot information.
In essence, DKF-DST leverages the coordinated function of its two primary stages to perform multi-domain dialogue state tracking, which not only strengthens its ability to manage dialogues across multiple domains but also enhances overall accuracy and performance.

\section{Experimental settings}

\subsection{Dataset}


In our study, we focus on the widely used MultiWOZ corpus, which remains one of the most influential benchmarks in the DST community. Dialogue state tracking research has benefited greatly from the emergence of MultiWOZ, a large and carefully constructed dataset targeting multi-domain conversational modeling. The dataset contains human–human dialogues and spans seven diverse domains (restaurant, hotel, attraction, taxi, hospital, police, and train). This broad coverage allows DST models to be evaluated in settings that closely resemble practical applications where users frequently switch between domains during a single conversation.

MultiWOZ consists of over ten thousand dialogues, with a significant portion involving complex multi-domain transitions. Such dialogues require models to maintain and update a rich set of slot value pairs as the user’s goals evolve, making the dataset particularly suitable for assessing a model’s ability to generalize across diverse task configurations. Each dialogue turn is annotated with complete dialogue states, enabling detailed supervision and facilitating both supervised learning and reinforcement learning approaches.

Because of its annotation quality, domain diversity, and scale, MultiWOZ has become the de facto standard for benchmarking DST systems. The dataset provides a rigorous testbed for evaluating dialogue models’ robustness, consistency, and cross-domain tracking capabilities, and continues to underpin much of the progress in contemporary dialogue state tracking research.

In the MultiWOZ dataset, “Wizard-of-Oz” refers to a data collection paradigm in which two participants asynchronously assume the roles of the user and the system, generating dialogue data based on a predefined task description. This setup helps increase the naturalness and diversity of the conversations. However, a major drawback of this crowdsourced collection process is the presence of inconsistent and inaccurate annotations. To mitigate these issues, four updated versions of the dataset were subsequently released, each aiming to correct the annotation errors:

MultiWOZ 2.1 \cite{eric2019multiwoz} serves as corrected version of the original dataset, addressing more than 32\% of state annotation errors across over 40\% of dialogue turns, as well as fixing 146 dialogue utterances. In addition, it incorporates user dialogue act information and provides multiple descriptions for each dialogue state slot, making it more suitable for low-resource, few-shot, and zero-shot learning scenarios.

MultiWOZ 2.2 \cite{zang2020multiwoz} enhances the earlier MultiWOZ 2.1 dataset by systematically identifying and correcting 17.3\% of the erroneous dialogue state annotations. It also provides a revised ontology and includes additional labels marking active intents and requested slots at every user turn, offering richer supervision for DST models.
MultiWOZ 2.3 \cite{han2021multiwoz}, while primarily dedicated to improving dialogue act annotations, leaves the noise and inconsistencies in dialogue state annotations largely untouched.

MultiWOZ 2.4 \cite{ye2021multiwoz} builds on MultiWOZ 2.1 with a clear objective of enhancing the robustness and fairness of dialogue state tracking evaluation. The update focuses on correcting inaccurate, inconsistent, and ambiguous annotations in the validation and test sets, ultimately revising more than 41\% of dialogue turns and over 65\% of session states. Although these substantial modifications significantly improve the reliability of evaluation, the training set remains unchanged. Empirical studies indicate that the refinements introduced in MultiWOZ 2.4 lead to consistently better performance than any earlier dataset version.

\subsection{Evaluation metrics}
Since what a system predicts as the dialogue state closely depends on the accumulated dialogue history, accuracy-focused metrics are widely used to assess the DST models. In this research, we primarily evaluate model effectiveness using two standard metrics: Joint Goal Accuracy (JGA), which measures correctness at the full-state level, and Slot Accuracy (SA), which assesses correctness at the individual slot level.

Joint Goal Accuracy (JGA) quantifies the proportion of dialogue turns in which the model’s predicted dialogue state is an exact match to the label dialogue state, thus serving as a rigorous measure of full-state prediction accuracy. Let $B_t$ and $B_t'$ denote the label and predicted dialogue state sets , respectively. The prediction for turn $t$ is considered correct only when $B_t'$ exactly matches $B_t$.

Unlike JGA, Slot Accuracy (SA) evaluates the correctness of the model’s predictions at the individual slot level for each dialogue turn. It is computed as the proportion of slot values that are predicted correctly, and is defined as:
\begin{equation}
SA=\frac{\sum_{i}^{n}acc_{i}}{n},
\end{equation}
where, $n$ denotes the number of slots in the ontology. To enable more detailed analysis, the evaluation can be further broken down by examining model performance across different slot categories.

Together, JGA and SA form the core evaluation metrics for dialogue state tracking. JGA assesses turn-level correctness of the complete dialogue state across the conversation, while SA isolates performance on each slot to reveal finer-grained accuracy patterns. Together, these metrics enable researchers and developers to assess and compare different DST models and offer valuable guidance for further model refinement and optimization.

\subsection{Baseline}
To evaluate the DKF-DST model, we adopt the following representative dialogue state tracking To thoroughly assess the effectiveness of DKF-DST, we compare it against several representative and widely adopted dialogue state tracking (DST) models:

TransformerDST \cite{zeng2020jointly}:
This method utilizes a pre-trained BERT backbone in combination with a Transformer encoder and adopts a two-stage prediction process for dialogue states. In the first stage, the model identifies whether each slot needs to be updated; in the second stage, it generates the corresponding slot value by leveraging contextual signals from the dialogue history.

SOM-DST \cite{kim2020efficient}:
SOM-DST adopts a selective state operation mechanism that determines the required update action for each slot before generating its value. Instead of relying on a heavy decoding process, the model directly replaces the previous slot value with the newly predicted one, thereby reducing computational overhead and improving overall inference efficiency.

TripPy \cite{heck2020trippy}:
TripPy incorporates three complementary copy mechanisms for populating slot values:
(1) Span prediction, which extracts value spans directly from user utterances;
(2) System-response copying, which captures values mentioned or confirmed by the system;
(3) Dialogue-state copying, which resolves intra- and inter-domain co-reference by copying existing values from the dialogue state.

SAVN \cite{wang2020slot}:
This model proposes an ontology-aware architecture based on Slot Attention (SA) and Value Normalization (VN). SA facilitates information sharing between slots and dialogue utterances, whereas VN enables span transformation and helps normalize diverse forms of slot values.

SimpleTOD \cite{hosseini2020simple}:
SimpleTOD employs a causal language model to jointly learn all sub-tasks of dialogue systems by casting them as a unified sequence-generation problem.

Seq2seq-DU \cite{seq2seq_DU}:
This approach reframes DST as a sequence generation task, making use of detailed BERT-based representations of both utterances and schema descriptions. The model supports flexible handling of categorical and non-categorical slot types and maintains strong adaptability when dealing with schema structures that were not observed during training.

D3ST \cite{zhao2022description}:
D3ST generates dialogue states solely from prompt-based descriptions by entirely replacing slot names or symbolic placeholders within the prompt. Additionally, an index-selection mechanism is incorporated to enhance flexibility and robustness in slot-value generation.

\subsection{Experiment settings}
The DKF-DST model is initialized with the publicly available pre-trained T5-base checkpoint \citep{raffel2020t5}, implemented via the HuggingFace Transformers framework \citep{wolf-etal-2020-transformers}. We employ a learning rate of 5e-5 and a batch size of 6, and optimize all trainable parameters using the AdamW optimizer \citep{DBLP:conf/iclr/LoshchilovH19}. All training and evaluation experiments are executed on a computing environment equipped with two NVIDIA RTX 3090 GPUs. Through extensive experimentation, we set the correlation score threshold in the first stage to 0.8.

\begin{table}[h]
\centering
\caption{Experimental results of the DKF-DST model}
\label{tab:main_result}
\begin{tabular}{ccccc}
\toprule
\textbf{model} & \textbf{MWZ2.1} & \textbf{MUZ2.2} & \textbf{MUZ2.3} & \textbf{MUZ2.4} \\
\midrule
Transformer-DST & 55.4 & - & - & - \\
SOM-DST&51.2  &-  &55.5 &66.8   \\
TripPy&55.3  &- &63.0 &59.6  \\
SAVN&54.5  &- &58.0 &60.1  \\
SimpleTOD&50.3  &- &51.3 &-  \\
Seq2seq&52.8  &57.6 &59.3 &67.1  \\

D3ST (Base)& 54.2 &56.1 &59.1 &72.1\\
D3ST (Large)& 54.5& 54.2 &58.6& 70.8\\
D3ST (XXL) &57.8 &58.7 &60.8 &75.9\\
\textbf{DKF-DST}& \textbf{58.2}& \textbf{62.3} &\textbf{63.1}& \textbf{77.3}\\

\bottomrule
\end{tabular}
\end{table}

\section{Experimental Results}\label{sec2}
\subsection{Comparison with baselines}

For the MultiWOZ 2.0–2.4 datasets, experiments were conducted to measure the predictive accuracy and overall performance of the DKF-DST model across different dataset versions and compare it with baseline models. The experimental outcomes are summarized in Table \ref{tab:main_result}, where “–” denotes unavailable public data, and bold values indicate the best performance.

Several recent baseline models were selected for comparison based on the following criteria:
(1) All baseline models focus on improving multi-domain dialogue state tracking (DST) performance and report experimental results trained on the full dataset.
(2) Each model adopts a sequence-to-sequence (seq2seq) training framework.
(3) Some baselines incorporate knowledge to enhance model performance—for example, the D3ST model effectively introduces slot ontology knowledge.

As shown in Table \ref{tab:main_result}, compared with other seq2seq-based methods for multi-domain DST, DKF-DST achieves the highest performance. Moreover, it can be observed that text-to-text models demonstrate significant improvements in DST tasks, particularly as the number of model parameters increases. Since the proposed model is trained using T5-XXL, results from D3ST under different model sizes are also listed for fair comparison. Overall, the proposed model outperforms D3ST in performance.

Compared with the D3ST model, the main advantage of this model lies in the innovative introduction of an information selection module, which enables the dynamic fusion of relevant slot information before performing state prediction. Unlike the D3ST model, which directly incorporates all slot information, our model adopts a more refined and efficient input design. By introducing the information selection module, the model effectively reduces input length and avoids introducing redundant information into the already lengthy dialogue context. This design not only improves model efficiency but also allows the system to handle information in a more flexible and intelligent manner, thereby enabling it to more accurately capture and predict dialogue states.

In addition, experimental results demonstrate that even when faced with the potential propagation of errors from the information selection stage to subsequent stages, the proposed model still achieves superior performance compared with the D3ST model. This indicates that our model possesses stronger robustness and stability in information selection and processing, effectively mitigating the challenges caused by error propagation. Through a well-designed information selection and fusion strategy, the model successfully overcomes the issues that may arise from propagated errors and ultimately achieves a higher level of performance than D3ST.

Overall, compared with the D3ST model, the proposed model demonstrates advantages in more precise input information processing, more efficient information fusion, and greater robustness and stability. These strengths enable the model to achieve superior performance in multi-domain dialogue state tracking tasks, showcasing enhanced modeling capability and prediction accuracy. Consequently, it provides an innovative and effective approach for addressing the complex challenges of dialogue state tracking in dialogue systems.

\subsection{Effects of parameter $\delta$}

\begin{table}[h]
\centering
\caption{The experiments with Different Thresholds}
\label{tab:Threshold}
\begin{tabular}{ccl}
\toprule
\textbf{$\delta$} & \textbf{Precision} & \textbf{Recall} \\
\midrule
0.9 & 86.2 & 98.4 \\
0.8& 96.8 &  98.1  \\
0.7& 85.7 & 98.6 \\
0.6& 78.3 & 99.5 \\
0.5& 52.3 & 99.8 \\
\bottomrule
\end{tabular}
\end{table}

Hyperparameter tuning is essential for achieving strong performance in our experiments. We focus on the role of the correlation score threshold in the first stage of DKF-DST. During this stage, a predefined threshold $\delta$ is used to discard irrelevant slots and retain those whose correlation scores surpass it, indicating relevance to the dialogue history. These filtered slots are subsequently used for knowledge fusion in stage two.
Recognizing that first-stage retrieval accuracy critically affects downstream outcomes, we conduct an empirical study on how different threshold settings of $\delta$ influence slot retrieval accuracy.

Specifically, in the experimental design, 20\% of the dataset was randomly sampled as experimental data to ensure the representativeness and reliability of the results. Then, different values were assigned to the correlation score threshold $\delta$ — for example, 0.9, 0.8, and 0.7 — to investigate how varying the threshold affects the slot retrieval accuracy in the first stage.
For each threshold value, the model’s performance metrics were recorded, with a primary focus on precision and other evaluation indicators, to rigorously evaluate the model’s performance under different threshold settings.

The precision and recall achieved under various threshold settings are summarized in Table \ref{tab:Threshold}. The analysis suggests that precision should serve as the primary evaluation metric for the first stage of identifying relevant slots. Precision highlights the proportion of true relevant slots among those predicted as relevant and thus directly measures the reliability of the model’s positive predictions. Since the core question in the first stage is whether the model can accurately identify the relevant slots from the dialogue history, precision provides a more meaningful assessment. Conversely, recall should not dominate the evaluation because it simply measures the fraction of actual relevant slots that are retrieved. As demonstrated in the experiments, lowering the threshold increases recall but simultaneously causes numerous irrelevant slots to be mislabeled as relevant, severely compromising precision.

In the first-stage task, greater emphasis is placed on the model’s accurate prediction of relevant slots, while a certain degree of deviation in failing to identify all relevant slots is acceptable. Therefore, by optimizing precision, the false positive rate can be effectively controlled, thereby improving the overall prediction accuracy of the model. In summary, although recall is also important in certain scenarios, precision is a more appropriate metric when evaluating the model’s accuracy in predicting relevant slots in the first stage. Emphasizing precision provides a clearer measure of the model’s ability to correctly identify relevant slots, improving slot extraction accuracy and strengthening overall model performance.

Table \ref{tab:Threshold} illustrates how precision varies with different threshold $\delta$ values, highlighting the importance of the relevance score threshold in the DKF-DST model. This analysis helps determine the optimal threshold for maximizing model performance. Empirical results indicate that a threshold $\delta$ of 0.8 yields the highest precision in first-stage relevant slot prediction. Accordingly, this value is employed as the relevance threshold in subsequent experimental inference. Exploring the influence of alternative threshold settings on performance allows for a more comprehensive understanding of the model’s inner workings and facilitates optimization of its slot retrieval accuracy and computational efficiency.

\subsection{Ablation analysis }

\begin{table}[h]
\centering
\caption{Ablation analysis}
\label{tab:Ablation_analysis}
\begin{tabular}{ccccc}
\toprule
\textbf{model} & \textbf{MWZ2.1} & \textbf{MWZ2.2} & \textbf{MWZ2.3} & \textbf{MWZ2.4} \\
\midrule
\textbf{DKF-DST}& \textbf{58.2}& \textbf{62.3} &\textbf{63.1}& \textbf{77.3}\\
- prompt& 45.6 &49.1& 52.1& 58.3 \\
- OT &47.8 &50.6 &53.6& 62.5\\
- CV &51.6 &55.8 &57.9& 63.7\\
\bottomrule
\end{tabular}
\end{table}

In the DKF-DST model, the design of prompts plays a crucial role in enhancing model performance.To further investigate the impact of prompting, we conducted an ablation study, as shown in Table \ref{tab:Ablation_analysis}, to examine the effect of the complete prompt as well as the contributions of its two main components (the output template and the candidate values). the content of the prompt significantly influences model performance. Without any prompt (“-prompt”), the model struggles to accurately understand and generate dialogue states. When prompts are introduced, however, the model becomes more adept at identifying the most relevant cues, enabling it to produce dialogue state outputs with higher precision.

Furthermore, the two components within the prompt — the output template (OT) and the candidate values (CV) — also play important roles. The output template provides guidance for dialogue behavior, helping the model generate responses that align with expectations. The candidate values supply predefined possible answers, enabling the model to select the most appropriate response.

The results demonstrate that both the prompt content and its two components are vital to the DKF-DST model’s performance. The careful design of prompts allows the model to focus more effectively on key information, thereby improving the accuracy of dialogue state prediction and the overall quality of generated outputs.

\section{Conclusion}\label{sec13}

In this work, we propose a dynamic knowledge fusion approach to tackle the inherent challenges in multi-domain dialogue state tracking. Multi-domain dialogues often involve complex interactions across different domains, which require models to effectively capture both dialogue history and contextual dependencies. To address these issues, we develop the Dynamic Knowledge Fusion for Multi-Domain Dialogue State Tracking (DKF-DST) model. Our approach leverages structured knowledge, including domain schemas and ontologies, to enrich the model’s understanding of dialogue context and slot relationships. By dynamically fusing this knowledge into the state tracking process, DKF-DST reduces model complexity, enhances generalization, and improves the overall accuracy of multi-domain dialogue state tracking, thereby strengthening the capability of dialogue systems in multi-domain settings.

\section*{Abbreviation}
\begin{tabularx}{\textwidth}{lX}
DST & Dialogue State Tracking \\
PTM & Pre-trained Model \\
LLM & Large Language Model \\
\end{tabularx}

\section*{Declarations}

\subsection*{Authors' contributions}
\textbf{Haoxiang Su}: Writing – original draft, collection and summary of references.
\textbf{Ruiyu Fang}: Summary of references.
\textbf{Liting Jiang}: Summary of references.
\textbf{Xiaomeng Huang}: Image.
\textbf{Shuangyong Song}: Discussion on technical solutions.

\subsection*{Competing Interest}
Not applicable.

\subsection*{Acknowledgements}
Not applicable.

\subsection*{Data availability}
Not applicable.

\subsection*{Code availability}
Not applicable.

\subsection*{Funding }
Not applicable.





\bibliography{sn-bibliography}

\end{document}